\documentclass[10pt,twocolumn,letterpaper]{article}

\usepackage{cvpr}  
\definecolor{cvprblue}{rgb}{0.21,0.49,0.74}
\usepackage[pagebackref,breaklinks,colorlinks,allcolors=cvprblue]{hyperref}
\usepackage{multirow}

\def\paperID{24} % *** Enter the Paper ID here
\def\confName{CVPR}
\def\confYear{2026}

\title{Sat3R: Satellite DSM Reconstruction via RPC-Aware Depth Fine-tuning}

\author{
Qiaoyi Yang$^{1}$,\quad
Chaoyi Zhou$^{1}$,\quad
Xi Liu$^{1}$,\quad
Run Wang$^{1}$,\quad
Minghui Xu$^{1}$,\quad
Mert D. Pes{\'e}$^{1}$\\
Feng Luo$^{1}$,\quad
Yuhao Xu$^{1}$,\quad
Zhi-Qi Cheng$^{2}$,\quad
Qiushi Chen$^{1}$,\quad
Hairong Qi$^{3}$,\quad
Siyu Huang$^{1}$\\[0.8em]
{\small
$^{1}$Clemson University
\qquad
$^{2}$University of Washington
\qquad
$^{3}$University of Tennessee
}
}

\begin{document}
\maketitle
\begin{abstract}
Accurate Digital Surface Model (DSM) reconstruction from satellite imagery 
is critical for applications such as disaster response, urban planning, and 
large-scale geographic mapping. Existing approaches face a fundamental 
trade-off: optimization-based methods achieve strong accuracy but require 
hours of per-scene computation, while generalizable geometry foundation 
models offer near-instant inference but fail to generalize to satellite 
imagery due to the domain gap introduced by the Rational Polynomial Camera 
(RPC) model and mismatched depth scale distributions. We present Sat3R, a 
feed-forward framework that bridges this gap via RPC-aware metric depth 
fine-tuning of Depth Anything V2 using the Scale-Invariant Logarithmic 
(SiLog) loss. By constructing physically consistent pseudo depth supervision 
from RPC geometry, Sat3R adapts a monocular depth foundation model to the 
satellite domain without per-scene optimization. Experiments on the DFC2019 
benchmark demonstrate that Sat3R reduces MAE by 38\% over zero-shot feed-forward baselines
and achieves competitive accuracy against optimization-based methods, while 
delivering over 300$\times$ speedup. Sat3R demonstrates that feed-forward models, when properly adapted to the 
satellite domain, can match optimization-based accuracy at a fraction of 
the computational cost, paving the way for practical large-scale satellite 
DSM reconstruction.

\begin{figure}[t]
    \centering
    \includegraphics[width=\linewidth]{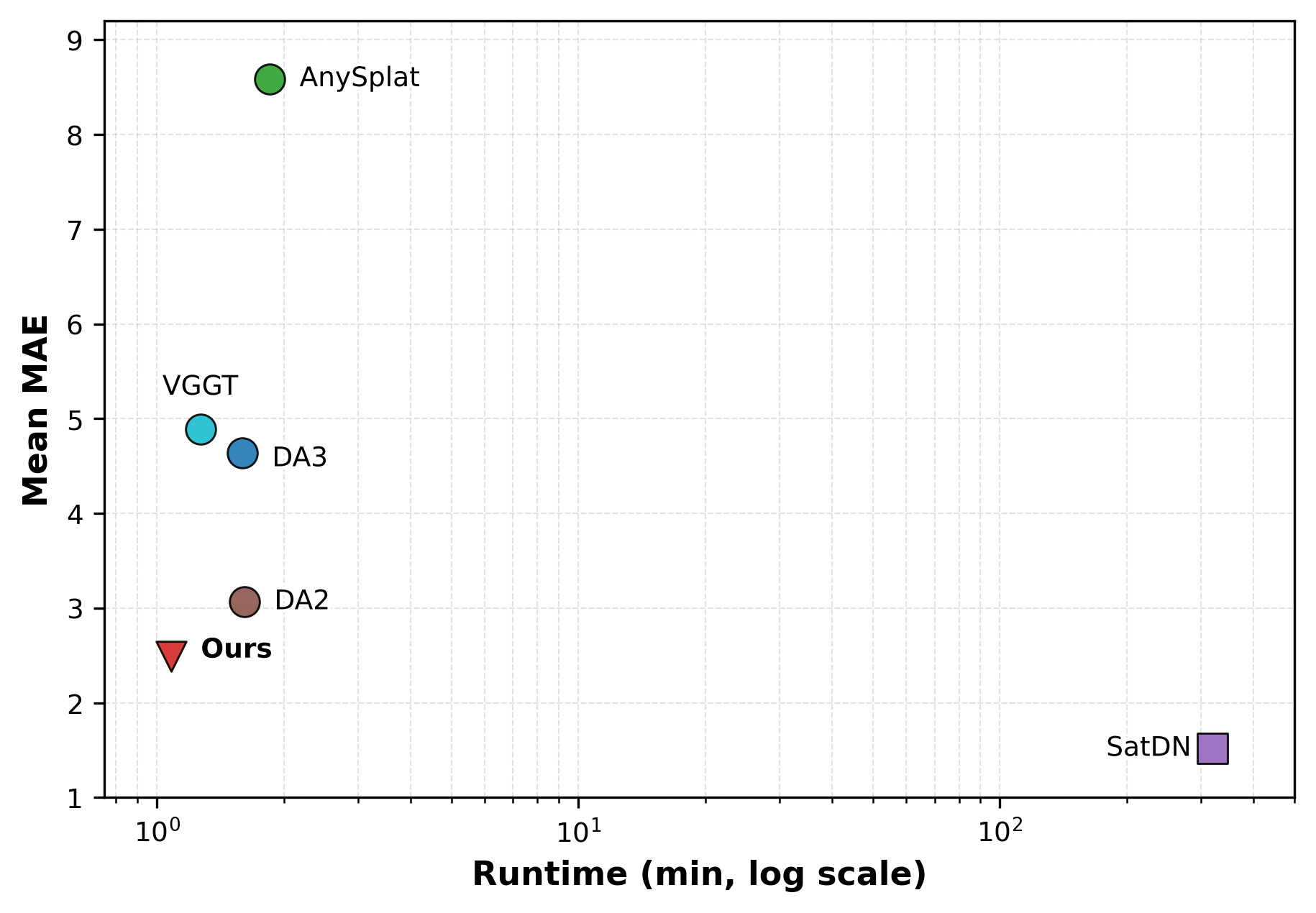}
    \vspace{-0.5cm}
\caption{Runtime vs.\ accuracy comparison on DFC2019~\cite{c6tm-vw12-19}. Sat3R is the first 
feed-forward framework for satellite DSM reconstruction, achieving 
comparable accuracy to optimization-based SatDN with over 300$\times$ 
speedup, while outperforming all zero-shot feed-forward baselines in 
reconstruction accuracy.}
    \label{fig:teaser}
    \vspace{-0.5cm}
\end{figure}

\end{abstract}    
\section{Introduction}
\label{sec:intro}

Accurate Digital Surface Model (DSM) reconstruction from satellite imagery 
supports a wide range of applications, including disaster response, urban 
planning, and large-scale geographic mapping. Given multi-view satellite 
images and their associated Rational Polynomial Camera (RPC) metadata, the 
goal is to recover a dense height map of the observed scene.

Existing approaches fall into two categories. Optimization-based methods 
such as SatDN~\cite{liu2025sat} achieve competitive DSM accuracy by fitting a neural 
representation per scene, but require hours of optimization per scene, 
making them impractical for time-sensitive applications. On the other hand, 
generalizable geometry foundation models (GFMs)~\cite{wang2025vggt,depth_anything_v2} offer near-instant 
inference without per-scene optimization, but perform poorly when applied 
directly to satellite imagery. As shown in Fig.~\ref{fig:teaser}, optimization-based methods achieve 
strong accuracy but at the cost of hours of computation per scene, while 
GFMs offer near-instant inference but with substantially degraded accuracy 
on satellite imagery --- no existing method delivers both the speed required 
for operational deployment and the accuracy demanded by downstream 
applications.

We identify two key reasons for this failure of GFMs on satellite data. 
First, satellite cameras follow the Rational Polynomial Camera (RPC) model, which has no explicit camera 
center, making the physical definition of depth fundamentally incompatible 
with the perspective-camera assumption underlying these models. Second, the 
depth scale distribution of satellite imagery differs substantially from the 
natural image datasets on which these models are trained.

To bridge this domain gap, we propose \textbf{Sat3R}, an RPC-aware fine-tuning approach. We construct image--depth training pairs from the training split of the DFC2019 satellite dataset ~\cite{c6tm-vw12-19}  by defining a physically consistent pseudo depth based on RPC geometry: for each pixel, we compute the slant-range distance between a near-plane reference point and the true surface 
intersection along the imaging ray. This pseudo depth serves as metric 
depth supervision to fine-tune Depth Anything V2 (DA2)~\cite{depth_anything_v2} using the 
Scale-Invariant Logarithmic (SiLog) loss~\cite{ZoeDepth}. At inference, predicted 
depths are back-projected through the RPC model to produce a DSM. As 
illustrated in Fig.~\ref{fig:teaser}, compared to zero-shot feed-forward baselines, our method reduces MAE by 
38\%. Compared to optimization-based SatDN, Sat3R achieves competitive 
accuracy while delivering over 300$\times$ speedup.

In summary, our contributions are:
\begin{itemize}
    \item We propose \textbf{Sat3R}, the first feed-forward framework for 
    satellite DSM reconstruction, enabling efficient inference without 
    per-scene optimization.
\item We introduce a pseudo depth construction pipeline for RPC satellite 
imagery, providing physically consistent metric depth supervision 
for fine-tuning monocular depth foundation models.
\item Extensive experiments on the DFC2019 dataset demonstrate that 
Sat3R achieves competitive DSM accuracy against optimization-based 
methods with over 300$\times$ speedup, while outperforming all zero-shot 
feed-forward baselines by 38\% in MAE.

\end{itemize}

\section{Related Work}
\label{sec:related_work}
\subsection{Optimization-Based 3D Reconstruction}
Classical multi-view stereo (MVS) methods reconstruct 3D structure by 
matching correspondences across images and optimizing for geometric 
consistency~\cite{schoenberger2016mvs,schoenberger2016sfm,schoenberger2016vote,pan2024glomap}. While effective, these pipelines require careful 
calibration and are computationally expensive at scale. Neural rendering 
approaches such as NeRF~\cite{mildenhall2020nerf,multinerf2022,wang2021neus,mueller2022instant} have significantly advanced reconstruction 
quality by representing scenes as continuous volumetric functions optimized 
per scene. More recent methods based on 3D Gaussian Splatting (3DGS)~\cite{kerbl3Dgaussians,Yu2024MipSplatting,liu20243dgs,zhou2025lrf,Huang2DGS2024} 
improve rendering efficiency while maintaining competitive reconstruction 
quality. However, the per-scene optimization nature of these methods results 
in inference times on the order of minutes to hours, limiting their 
applicability in time-sensitive scenarios such as disaster response and 
large-scale mapping.
\subsection{Generalizable Geometry Foundation Models}
Recent work has demonstrated that large-scale pretraining can yield powerful geometry estimators~\cite{wang2025vggt,dust3r_cvpr24,zhang2024monst3r,mast3r_eccv24,wang2025continuous,depth_anything_v2,wang2026flexmapgeneralizedhdmap,Yang_2025_Fast3R,jiang2025anysplat,depthanything3,zhou2026ff3r} that generalize across scenes without per-scene optimization. For monocular depth estimation, Depth Anything (DA)~\cite{depthanything,depth_anything_v2,depthanything3} achieves strong zero-shot performance on natural images by training on large-scale labeled and unlabeled data. For multi-view geometry, models such as VGGT~\cite{wang2025vggt} and AnySplat~\cite{jiang2025anysplat} directly infer 3D structure from multiple input views in a single feed-forward pass. Despite their efficiency, these models are trained exclusively on natural image datasets and fail to generalize to satellite imagery, where the RPC camera model and depth scale distribution differ fundamentally from the perspective-camera assumption underlying these methods. This domain gap motivates our work, which constructs RPC-aware training 
data to adapt DA2~\cite{depth_anything_v2} for satellite DSM reconstruction.
\subsection{Satellite 3D Reconstruction}

Early attempts to apply neural rendering to satellite imagery adapt NeRF 
to the unique properties of satellite acquisition. Several works focus on 
appearance modeling, incorporating shadow 
cues~\cite{mari2022sat}, and dynamic object uncertainty~\cite{Mari_2023_CVPR} to improve novel view 
synthesis quality. Others target geometric accuracy, introducing 
photo-consistency constraints~\cite{Sat_Mesh} or hash-grid representations~\cite{billouard2024satngp} 
to accelerate optimization. SatDN~\cite{liu2025sat} extends this line by adapting the 
RPC camera model into a NeRF-based framework, achieving strong DSM accuracy 
but requiring hours of per-scene optimization. Our work departs from this 
paradigm by adopting a feed-forward approach, eliminating per-scene 
optimization entirely.
\section{Method}
\label{method}

\begin{figure*}[t]
\vspace{-0.5cm}
    \centering
    \includegraphics[width=\linewidth]{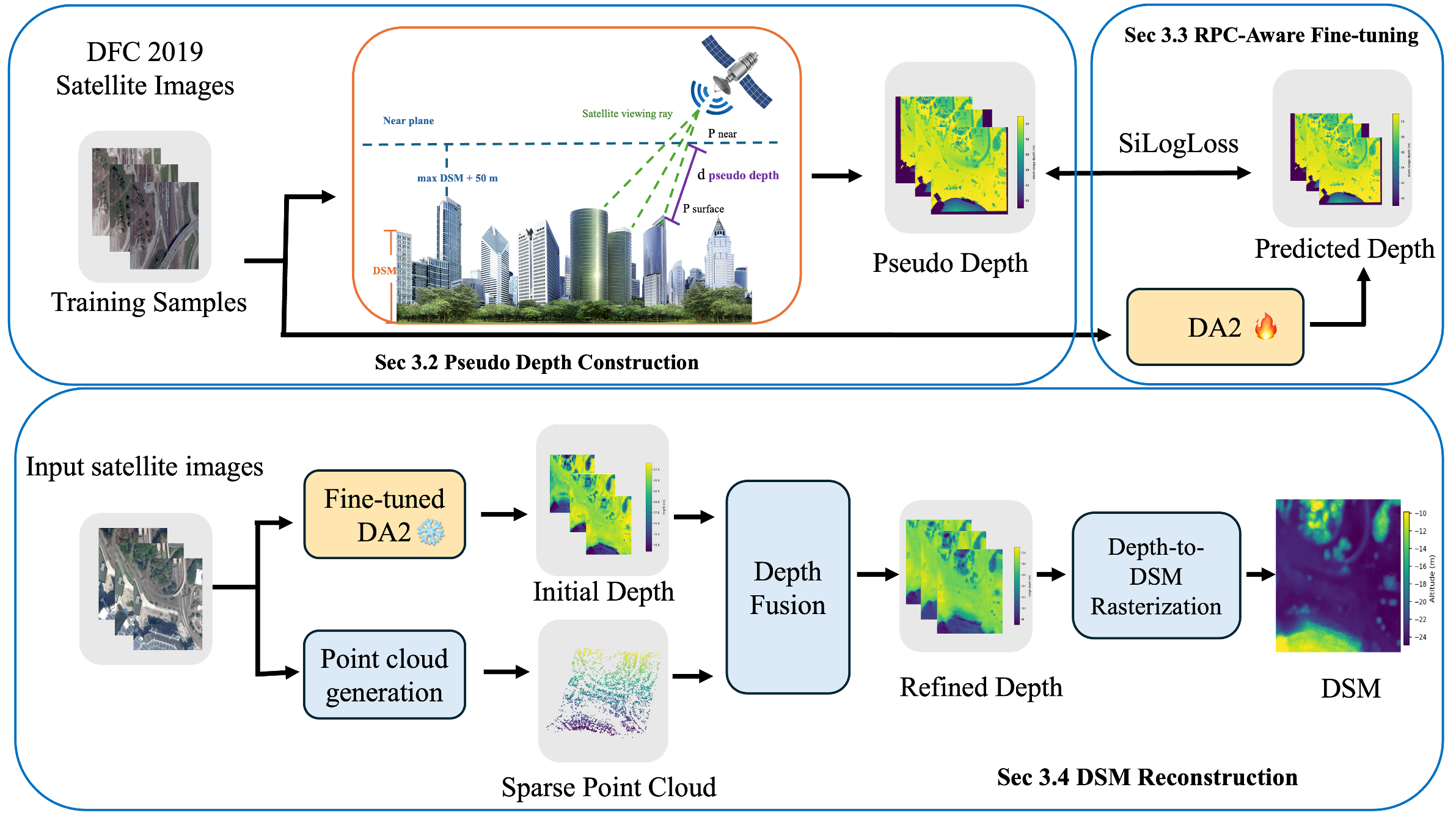}
    \vspace{-0.5cm}
    \caption{Overview of Sat3R. Given multi-view satellite images and their 
    associated RPC models, we first construct pseudo depth supervision via 
    slant-range geometry. The constructed 
    image--depth pairs are used to fine-tune DA2 in an RPC-aware manner. At inference, per-view depth maps are fused 
    and back-projected through the RPC model to produce the final DSM.}
    \vspace{-0.5cm}
    \label{fig:method}
\end{figure*}
\subsection{Problem Formulation}
Given a set of $N$ overlapping satellite images $\{I_i\}_{i=1}^{N}$ and 
their associated RPC models $\{R_i\}_{i=1}^{N}$, our goal is to reconstruct 
a Digital Surface Model (DSM) representing the height of the scene at each 
geographic location. The overall pipeline of Sat3R is shown in Fig.~\ref{fig:method}.

\subsection{Pseudo Depth Construction}
Satellite imagery is governed by the RPC model, which provides no explicit 
camera center, making standard perspective depth undefined. We define a 
physically consistent pseudo depth analogous to the camera-to-surface 
distance in perspective geometry.

For each image, we define a near-plane at altitude:
\begin{equation}
    z_{\text{ref}} = z_{\text{max}} + \delta
\end{equation}
where $z_{\text{max}}$ is the maximum scene altitude from image metadata 
and $\delta = 50$\,m is a fixed margin. For each pixel, we compute its RPC 
imaging ray and find its intersection with the near-plane 
$\mathbf{p}_{\text{near}}$ and with the true surface $\mathbf{p}_{\text{surface}}$ 
via fixed-point iteration on the ground-truth DSM. The pseudo depth is then 
defined as the slant-range distance:
\begin{equation}
    d = \|\mathbf{p}_{\text{surface}} - \mathbf{p}_{\text{near}}\|_2
\end{equation}
This construction provides per-pixel metric depth supervision for DA2~\cite{depth_anything_v2}. Training pairs are curated from DFC2019~\cite{c6tm-vw12-19} 
by retaining scenes with complete RGB--CLS alignment, sufficient multi-view 
coverage, and excluding winter scenes identified via IMD acquisition 
timestamps. 

\subsection{RPC-Aware Fine-tuning}
We fine-tune DA2~\cite{depth_anything_v2} on the constructed image--depth pairs following 
the metric depth fine-tuning protocol of DA2, using the Scale-Invariant 
Logarithmic (SiLog) loss~\cite{ZoeDepth}:
\begin{equation}
    \resizebox{0.85\columnwidth}{!}{$
    \mathcal{L}_{\text{SiLog}} = \dfrac{1}{N}\sum_{i=1}^{N} 
    \left(\log \hat{d}_i - \log d_i\right)^2 - 
    \dfrac{\lambda}{N^2}\left(\sum_{i=1}^{N} 
    \log \hat{d}_i - \log d_i\right)^2
    $}
\end{equation}
where $\hat{d}_i$ and $d_i$ are the predicted and pseudo ground-truth 
depth values respectively, and $\lambda$ is a variance minimization 
factor. The maximum depth range for inference is set to 150\,m to align with 
the satellite scene depth distribution. This enables Sat3R to directly output metric-scale depth without 
requiring scale recovery in post-processing. 

\subsection{DSM Reconstruction}
At inference, DA2 predicts per-view depth maps with a maximum depth 
range of 150\,m, calibrated to the satellite scene depth distribution.
Per-view depth maps are fused into metric depth using the multi-view fusion 
module~\cite{liu2025sat}. Each pixel's fused depth  $d^*_i$ is back-projected 
into 3D space along its RPC imaging ray:
\begin{equation}
    \mathbf{P}_i = \text{RPC}^{-1}(u_i, v_i, d^*_i)
\end{equation}
where $(u_i, v_i)$ is the image coordinate and $\mathbf{P}_i \in \mathbb{R}^3$ 
is the recovered 3D point in geographic coordinates (longitude, latitude, 
altitude). The 3D points from all views are then aggregated and rasterized 
onto a uniform 2D grid, with height values pooled via p90 to produce the 
final DSM $\mathbf{H} \in \mathbb{R}^{W \times H}$. Since RPC-aware fine-tuning outputs 
metric-scale depth, it provides a better initial estimate for the fusion 
stage, accelerating convergence of the depth fusion optimization.

% \subsection{DSM Reconstruction}

% At inference, the fine-tuned DA2 predicts per-view metric depth maps with a
% maximum depth range of 150\,m. These depths are fused using the multi-view
% fusion module~\cite{liu2025sat} to obtain refined metric depths.

% For each valid pixel $(u_i, v_i)$, we recover its RPC imaging ray by
% intersecting the RPC model with altitude planes. Let
% $\mathbf{p}^{\mathrm{near}}_i$ be the near-plane intersection and
% $\mathbf{r}_i$ be the unit ray direction. The fused depth $d_i^*$ is treated
% as the slant-range distance from the near plane, and the 3D point is computed as
% \begin{equation}
%     \mathbf{P}_i =
%     \mathbf{p}^{\mathrm{near}}_i + d_i^* \mathbf{r}_i .
% \end{equation}

% The reconstructed 3D points from all views are rasterized onto a uniform DSM
% grid, where height values within each cell are aggregated using p90 pooling to
% produce the final DSM $\mathbf{H} \in \mathbb{R}^{W \times H}$.
\section{Experiments}
\label{sec:exp}

\begin{table*}[t]
\centering
\resizebox{\textwidth}{!}{
\small
\setlength{\tabcolsep}{4.2pt}
\renewcommand{\arraystretch}{1.08}
\begin{tabular}{lcccccccc}
\toprule
\textbf{Method}
& \textbf{JAX\_207}
& \textbf{JAX\_214}
& \textbf{JAX\_260}
& \textbf{OMA\_212}
& \textbf{OMA\_287}
& \textbf{OMA\_315}
& \textbf{Mean}
& \textbf{Time$\downarrow$} \\
& MAE$\downarrow$/MED$\downarrow$
& MAE$\downarrow$/MED$\downarrow$ 
& MAE$\downarrow$/MED$\downarrow$
& MAE$\downarrow$/MED$\downarrow$
& MAE$\downarrow$/MED$\downarrow$
& MAE$\downarrow$/MED$\downarrow$
& MAE$\downarrow$/MED$\downarrow$
&  \\
\midrule
\multicolumn{9}{l}{\textit{Optimization-based methods}} \\
SatDN~\cite{liu2025sat}
& \textbf{2.399} / \textbf{1.214}
& \textbf{1.816} / \textbf{0.732}
& \textbf{1.620} / \textbf{0.961}
& \underline{0.888} / \underline{0.529}
& \textbf{0.894} / \textbf{0.405}
& \textbf{1.022} / \textbf{0.598}
& \textbf{1.44} / \textbf{0.74}
& 5h 36min \\
\midrule
\multicolumn{9}{l}{\textit{Feed-forward methods}} \\
AnySplat~\cite{jiang2025anysplat}
& 4.477 / 3.743
& 8.689 / 6.665
& 4.897 / 4.138
& 1.123 / 1.938
& 3.153 / 2.247
& 2.930 / 2.201
& 4.21 / 3.49
& 1min 37s \\
VGGT~\cite{wang2025vggt}
& 5.519 / 4.476
& 6.346 / 4.121
& 3.699 / 2.711
& 2.786 / 2.234
& 3.804 / 2.509
& 2.483 / 1.647
& 4.11 / 2.95
& 1min 31s \\
DA3~\cite{depthanything3}
& 4.647 / 3.653
& 8.341 / 5.651
& 4.658 / 5.651
& 2.118 / 1.193
& 3.557 / 2.816
& 2.887 / 2.058
& 4.37 / 3.50
& 1min 40s \\
DA2~\cite{depth_anything_v2}
& 4.578 / 3.254
& 7.841 / 5.403
& 4.014 / 3.224
& 4.230 / 3.078
& 3.681 / 3.203
& 3.214 / 2.486
& 4.59 / 3.44
& \underline{1min 26s} \\
\textbf{Ours}
& \underline{3.437} / \underline{2.082}
& \underline{5.372} / \underline{2.934}
& \underline{3.515} / \underline{2.573}
& \textbf{0.781} / \textbf{0.446}
& \underline{2.528} / \underline{1.974}
& \underline{1.305} / \underline{0.662}
& \underline{2.82} / \underline{1.78}
& \textbf{1min 05s} \\
\bottomrule
\end{tabular}
}

\caption{DSM reconstruction accuracy on six scenes from DFC2019 dataset~\cite{c6tm-vw12-19}. Ours achieves the best performance among feed-forward methods and approaches optimization-based SatDN, while running over 300$\times$ faster.}
\label{tab:selected_six_scenes_runtime}
\end{table*}

\begin{figure*}[t]
    \centering
    \includegraphics[width=0.95\textwidth]{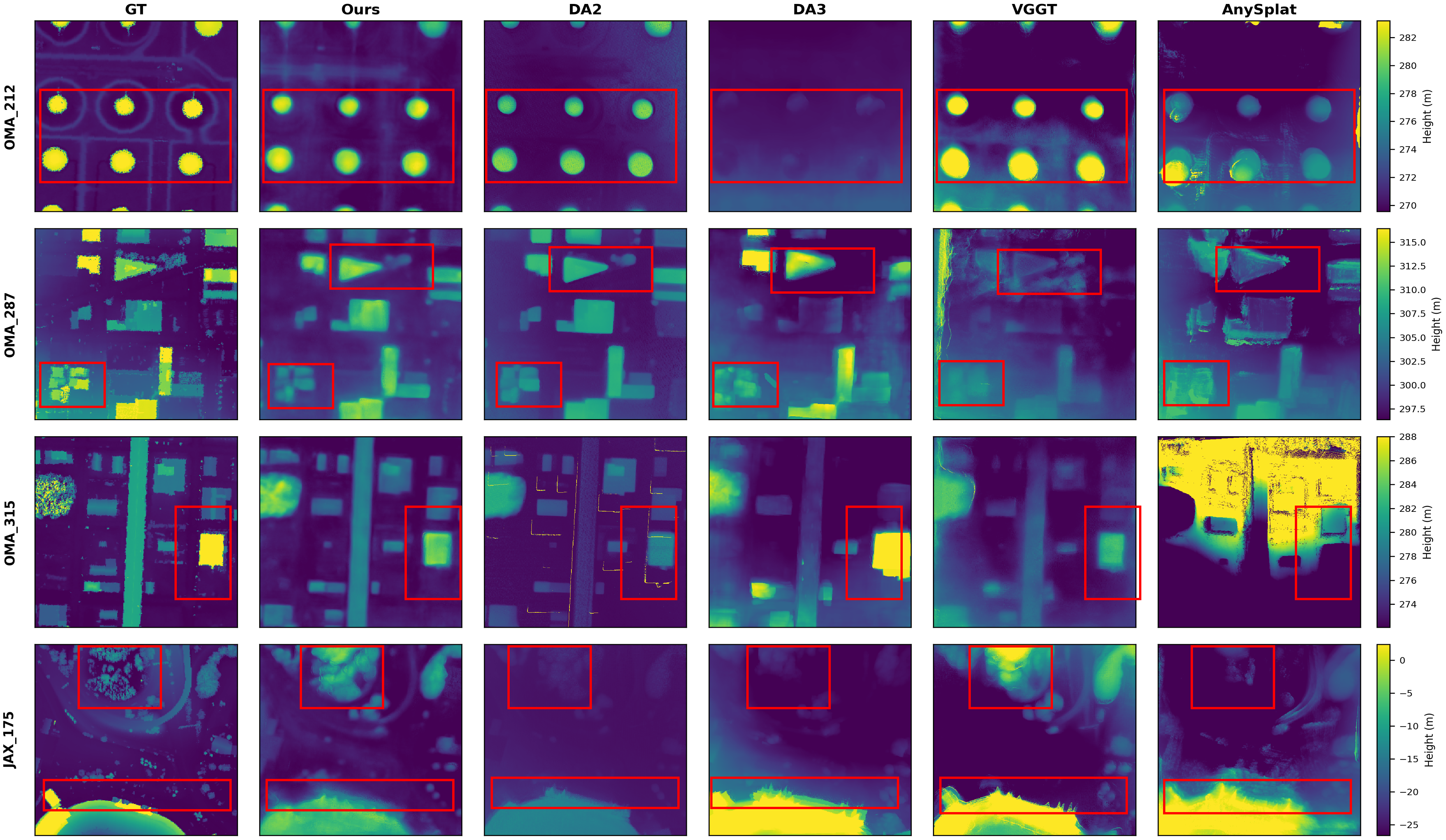}
    \caption{Qualitative comparison of DSM reconstruction results on selected DFC2019 scenes. 
Red boxes highlight representative regions where different methods show visible structural differences.
}
\vspace{-0.5cm}
    \label{fig:qualitative}

\end{figure*}

\subsection{Implementation Details}
We train and evaluate on the DFC2019 satellite dataset~\cite{c6tm-vw12-19}. Please refer to the appendix~\ref{sec:app_impl} for more details. 

\subsection{Experiment Results}

As shown in Table~\ref{tab:selected_six_scenes_runtime}, Sat3R consistently outperforms all feed-forward baselines, reducing mean MAE from 4.59 for zero-shot DA2 to 2.82 and mean MED from 3.44 to 1.78, 
demonstrating that RPC-aware metric depth fine-tuning effectively bridges 
the domain gap between natural and satellite imagery. Despite being a 
feed-forward method, Sat3R closely approaches the accuracy of the 
optimization-based upper bound SatDN, while delivering over 300$\times$ 
speedup.

As shown in Fig.~\ref{fig:qualitative}, zero-shot feed-forward baselines 
suffer from severe scale drift and blurry boundaries due to the domain gap 
between natural and satellite imagery. In contrast, Sat3R produces DSMs 
with accurate scale range, sharper building boundaries, and better 
preservation of fine-grained geometric details, closely matching the 
ground truth.

\vspace{-0.2cm}

\subsection{Ablation Study}
\vspace{-0.2cm}
% \begin{table}[t]
% \centering
% \small
% \setlength{\tabcolsep}{6pt}
% \renewcommand{\arraystretch}{1.08}
% \begin{tabular}{lcccc}
% \toprule
% \textbf{Method}
% & \textbf{Max Depth}
% & \textbf{Mean MAE}$\downarrow$
% & \textbf{Mean MED}$\downarrow$ \\
% \midrule
% DA2-FTD
% & 100
% & 3.312
% & 2.254 \\
% \textbf{DA2-FTD}
% & \textbf{150}
% & \textbf{3.131}
% & \textbf{1.963}\\
% DA2-FTD
% & 300
% & 3.412
% & 2.354 \\
% \bottomrule
% \end{tabular}
% \caption{
% Ablation study on the maximum depth threshold.
% }
% \label{tab:ablation_max_depth}
% \end{table}

We investigate the effect of the maximum depth threshold during inference. Detailed ablation results are provided in Appendix~\ref{sec:app_ablation}.

\section{Conclusion}
\vspace{-0.1cm}
\label{sec:con}
We present Sat3R, a feed-forward framework for satellite DSM reconstruction 
that bridges the domain gap between natural and satellite imagery via 
RPC-aware fine-tuning of Depth Anything V2. By constructing physically 
consistent pseudo depth supervision from RPC geometry, Sat3R adapts a 
monocular depth foundation model to the satellite domain without per-scene 
optimization. Experiments on DFC2019~\cite{c6tm-vw12-19} demonstrate that Sat3R achieves 
competitive accuracy against optimization-based methods while delivering 
over 300$\times$ speedup, establishing a practical and efficient baseline 
for satellite DSM reconstruction.
{
    \small
    \bibliographystyle{ieeenat_fullname}
    \bibliography{main}
}

\clearpage
\appendix

\newpage
\section*{Appendix}
\noindent In the Appendix, we provide the following: 
\begin{itemize}
    \item comprehensive implementation details in Section~\ref{sec:app_impl}
    \item comprehensive ablation details in Section~\ref{sec:app_ablation}
\end{itemize}
\section{Implementation Details}
\label{sec:app_impl}
Training pairs are constructed from scenes with complete RGB--CLS alignment 
and sufficient multi-view coverage, with winter scenes excluded via IMD 
acquisition timestamps. We fine-tune DA2~\cite{depth_anything_v2} for 40 epochs with a learning 
rate of $5\times10^{-6}$ using the AdamW optimizer on 2$\times$NVIDIA A100 
40GB GPUs. All inference experiments are conducted on a single NVIDIA A100 
40GB GPU. At inference, per-view depth maps are fused using 
the SatDN fusion module~\cite{liu2025sat} and rasterized to a DSM grid via p90 pooling. 
We evaluate on 6 held-out scenes (JAX\_207, JAX\_214, JAX\_260, OMA\_212, 
OMA\_287, OMA\_315) unseen during training, using Mean Absolute Error (MAE) 
and Median Error (MED) as metrics.

We compare against two categories of methods. 
\textbf{Optimization-based:} SatDN~\cite{liu2025sat}, which fits a neural 
representation per scene and serves as the accuracy upper bound. 
\textbf{Feed-forward:} DA2~\cite{depth_anything_v2}, DA3~\cite{depthanything3}, VGGT~\cite{wang2025vggt}, and 
AnySplat~\cite{jiang2025anysplat}. For all feed-forward baselines, per-view depths are processed through 
the same depth fusion, alignment, and rasterization pipeline to ensure 
a fair comparison.

\section{Ablation Study}
\label{sec:app_ablation}
\begin{table}[t]
\centering
\small
\setlength{\tabcolsep}{6pt}
\renewcommand{\arraystretch}{1.08}
\begin{tabular}{ccc}
\toprule
\textbf{Max Depth}
& \textbf{Mean MAE}$\downarrow$
& \textbf{Mean MED}$\downarrow$ \\
\midrule
100
& 3.312
& 2.254 \\
300
& 3.412
& 2.354 \\
\textbf{150}
& \textbf{3.131}
& \textbf{1.963} \\
\bottomrule
\end{tabular}
\caption{
Ablation study on the maximum depth threshold.
}
\label{tab:ablation_max_depth}
\end{table}

\begin{figure}[t]
    \centering
    \includegraphics[width=\linewidth]{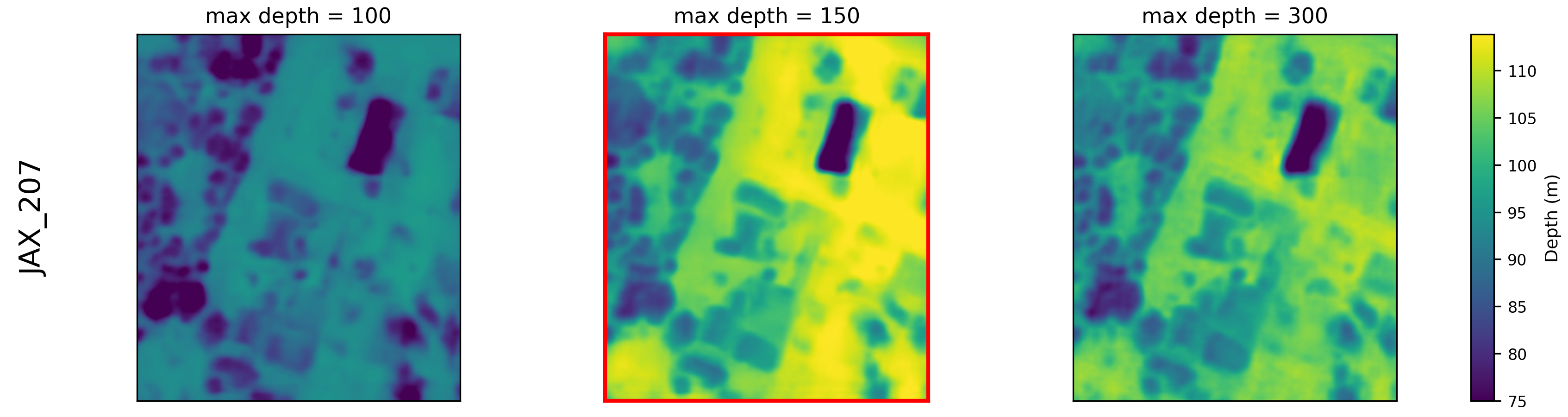}
    
    \vspace{1mm}
    
    \includegraphics[width=\linewidth]{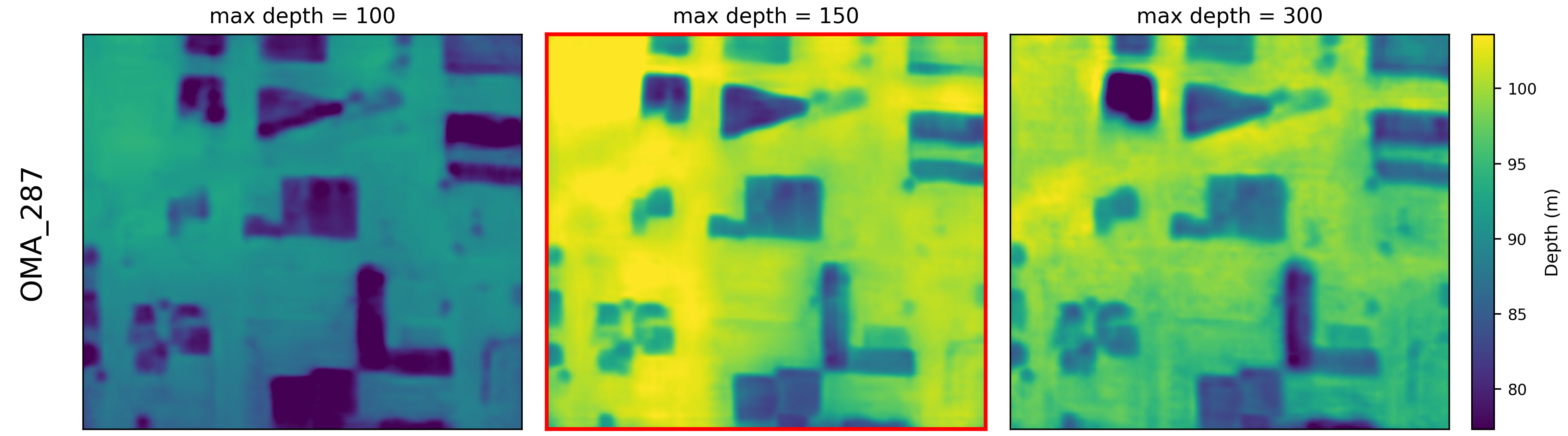}
    
    \caption{
    Qualitative ablation on the maximum depth threshold. 
    Setting the threshold to 150m produces sharper boundaries and more accurate height estimation compared to 100m and 300m.
    }
    \label{fig:max_depth_ablation_vis}
\end{figure}

As shown in Table~\ref{tab:ablation_max_depth} and Fig.~\ref{fig:max_depth_ablation_vis}, a threshold of 150m achieves the 
best performance across all metrics. A smaller value (100m) causes 
saturation near the depth boundary, while a larger value (300m) reduces 
effective depth resolution relative to the actual scene range 
($\sim$55--94\,m), degrading accuracy. This confirms that aligning the 
depth range with the satellite data distribution is critical for effective 
fine-tuning.

\end{document}